\documentclass[11pt,a4paper]{article}
\usepackage{CJKutf8}
\usepackage{graphicx} 
\usepackage{float} 
\usepackage{subfigure} 
\PassOptionsToPackage{hyphens}{url}
\usepackage[hyperref]{eacl2021}
\usepackage{times}
\usepackage{latexsym}

\usepackage{linguex}
\usepackage{booktabs}

\usepackage{microtype}

\aclfinalcopy 

\title{CLiMP: A Benchmark for Chinese Language Model Evaluation}
  
\author{\parbox{\linewidth}{\centering
Beilei Xiang,{\rm\affmark[1]} Changbing Yang,{\rm\affmark[1]} Yu Li,{\rm\affmark[1]} Alex Warstadt{\rm\affmark[2]} \and Katharina Kann{\rm\affmark[1]}} \vspace{.12cm}
\\
\affaddr{\affmark[1]University of Colorado Boulder},
\affaddr{\affmark[2]New York University}\\
\affaddr{\texttt{\{beilei.xiang, changbing.yang, yuli9309\}@colorado.edu}} \\
\affaddr{\texttt{warstadt@nyu.edu}}\\
\affaddr{\texttt{katharina.kann@colorado.edu}}}
\newcommand*{\affaddr}[1]{#1} 
\newcommand*{\affmark}[1][*]{\textsuperscript{#1}}

\date{}

\begin{document}
\begin{CJK}{UTF8}{gbsn}
\maketitle
\begin{abstract}
Linguistically informed analyses of language models (LMs) contribute to the understanding and improvement of these models. Here, we introduce the corpus of Chinese linguistic minimal pairs (CLiMP), which can be used to investigate what knowledge Chinese LMs acquire. CLiMP consists of sets of 1,000 minimal pairs (MPs) for 16 syntactic contrasts in Mandarin, covering 9 major Mandarin linguistic phenomena. 
The MPs are semi-automatically generated, and human agreement with the labels in CLiMP is $95.8\%$. 
We evaluate 11 different LMs on CLiMP, covering \emph{n}-grams, LSTMs, and Chinese BERT. 
We find that classifier–noun agreement and verb complement selection are the phenomena that models generally perform best at. However, models struggle the most with the \emph{b\v{a}} construction, binding, and filler-gap dependencies. Overall, Chinese BERT achieves an $81.8\%$ average accuracy, while the performances of LSTMs and 5-grams are only moderately above chance level. 

\end{abstract}

\section{Introduction}
Language models (LMs) are crucial parts of natural language processing (NLP) systems for a large variety of tasks, including summarization, machine translation, and dialog generation. More recently, they have become popular in the form of pretrained models,\footnote{Throughout this paper, we adopt a broad definition of LMs, which includes language representation models which have been trained on a masked language modeling objective.}  which are then fine-tuned on downstream tasks and often obtain state-of-the-art performance \cite{peters-etal-2018-deep,devlin-etal-2019-bert,conneau-etal-2020-unsupervised}. However, which linguistic phenomena language models can or cannot learn is still poorly understood for many languages.

Resources for the syntactic evaluation of LMs, such as BLiMP \cite{TACL2013} have focused mainly on English, and non-English resources currently only cover a small set of phenomena  \citep{mueller2020cross,gulordava-etal-2018-colorless, ravfogel-etal-2018-lstm}. 
In order to spur the analysis and subsequent improvement of LMs in Chinese, we introduce the corpus of Chinese linguistic minimal pairs (CLiMP), which can be used to evaluate LMs' knowledge of Chinese grammar. 

CLiMP consists of 16 individual datasets that are semi-automatically generated from grammar templates. Each set---or \textit{paradigm}---contains 1,000 minimal pairs (MPs). Together, they cover 9 core linguistic phenomena in Chinese. Human agreement on this corpus is $95.8\%$, confirming that CLiMP represents robust contrasts in Chinese grammar. High performance on CLiMP thus implies high correlation with human acceptability judgments across these phenomena.

We use CLiMP to study Chinese BERT \cite{devlin-etal-2019-bert},\footnote{\href{https://github.com/google-research/bert/blob/master/multilingual.md}{https://github.com/google-research/bert/blob/master/multilingual.md}} 6 LSTM \citep{hochreiter1997long} LMs, and 4 \emph{5}-gram LMs. We evaluate for each MP whether the LM assigns a higher probability to the grammatical or the ungrammatical sentence. 
Our results show that Chinese BERT is closest to human performance, achieving an 81.8\% accuracy on average over all phenomena, while the performances of LSTMs and \emph{5}-grams, regardless of the training data size, are only moderately above chance level.
Classifier--noun agreement and verb complement selection are the phenomena that models generally perform best at, suggesting that Chinese LMs are better at acquiring 
knowledge of local selectional restrictions. The \emph{b\v{a}} construction, binding, and filler-gap dependencies are the phenomena models have the most difficulties with. This indicates that they struggle to learn hierarchical syntax and to identify long-distance dependencies.


\section{Related Work}
\subsection{Language Models}
LMs assign probabilities to sequences of words \citep{Jurafsky2009}. Recently, they have become commonly used as pretrained models, which can be fine-tuned for downstream NLP tasks \citep{peters-etal-2018-deep,devlin-etal-2019-bert,conneau-etal-2020-unsupervised}.
Strictly speaking, LMs compute the probabilities of words based only on past context. BERT \cite{devlin-etal-2019-bert}, however, is trained using a masked language modeling objective: it predicts words based on past and future tokens. \citet{wang-cho-2019-bert} show that BERT is a Markov random field language model that can assign sentences a pseudo-log-likelihood score, which is computed by summing the conditional log
probabilities 
of all tokens in the sentence, as well as generate text. \citet{shin2019effective} and \citet{salazar-etal-2020-masked} apply pseudo-log-likelihood scores to sentence ranking and LM evaluation.

\subsection{Evaluation of Linguistic Knowledge}
Numerous methods exist for probing syntactic knowledge of neural network models in English \citep{hewitt-manning-2019-structural,tenney2019you}, and a growing body of work evaluates the syntactic knowledge of neural models by testing whether they can judge the grammatical acceptability of sentences. One common version of this task uses MPs to evaluate LMs' linguistic knowledge  \citep{Linzen2016AssessingTA,marvin-linzen-2018-targeted,TACL2013, wilcox2018rnn}. 

A MP is a pair of sentences that only differ in acceptability due to a single edit, as in (1) and (2). Native speakers can be asked to choose which sentence in each pair sounds more grammatical. 
Semi-automatically generating MPs can yield a larger set of controlled sentences, providing sufficient data for model evaluation \citep{Linzen2016AssessingTA,marvin-linzen-2018-targeted, ettinger-etal-2018-assessing}.

\exg. 王鑫 \underline{把}   自行车 扔 了。 \\
Wángxīn \underline{bǎ} zìxíngchē rēng le\\
SUBJ. \ \underline{BA.}   OBJ. \ \quad V. \ PST.\\
``Xin Wang threw away a bike."

\exg. 王鑫  \underline{被}  自行车 扔 了。 \\
Wángxīn \underline{bèi} zìxíngchē rēng le
\\
SUBJ.  \underline{PASS.}   OBJ.  \quad V. \ PST.\\
``Xin Wang was thrown away by a bike."

It is possible to model acceptability in a totally unsupervised way using LMs. The model assigns a probability to each sentence in a MP, and the one with the higher score is predicted as correct, and the model's predictions can be evaluated against human judgments  \citep{marvin-linzen-2018-targeted,TACL2013}. Supervised approaches are also possible \citep{warstadt2019neural}, but can be less informative on LMs' linguistic knowledge acquisition due to the bias introduced by training on acceptability judgment labels. 

Some prior work evaluates the linguistic knowledge of different non-English models \cite{ravfogel-etal-2018-lstm,gulordava-etal-2018-colorless,mueller2020cross}. However, these efforts focus mainly on subject-verb agreement, which is absent in Chinese, and the knowledge of Chinese LMs has not yet been explicitly studied.

Finally, the linguistic abilities of English BERT have been investigated in a a lot of prior work, e.g., \citet{clark2019does,vig2019visualizing,hewitt-manning-2019-structural}. We refer the reader to  \citet{rogers2020primer} for an overview.

\section{CLiMP}
Our main contribution is CLiMP, a corpus of Chinese MPs designed to evaluate Chinese LMs. CLiMP consists of 1,000 MPs for each of 16 grammatical contrasts, covering 9 major Chinese linguistic phenomena. Example MPs for each phenomenon are shown in Table \ref{MP-examples}.

\subsection{Data Generation}      
We generate data from grammar templates for every paradigm we incorporate. Our templates set lexical, syntactic, and semantic constraints for each paradigm, aiming at building robust contrasts and keeping the sentence length the same within each MP. 
We then build an annotated vocabulary, and generate sentences by sampling words from it. (1) and (2) show an MP  together with the template\footnote{The template example is only for demonstrative purposes. More information is encoded for the actual data generation.} used to create it. 

\subsection{Vocabulary}
We translate \citeauthor{TACL2013}'s (\citeyear{TACL2013}) English vocabulary, containing 3,000 English words with morphological, syntactical, and semantic annotations. We add words and features specific to Chinese linguistic phenomena to our vocabulary, including classifiers, verb complements, action verbs, and coverbs. Our final vocabulary contains 3,456 words and 84 features. 

\begin{figure}[h!]
     \centering
     \includegraphics[width=0.47\textwidth]{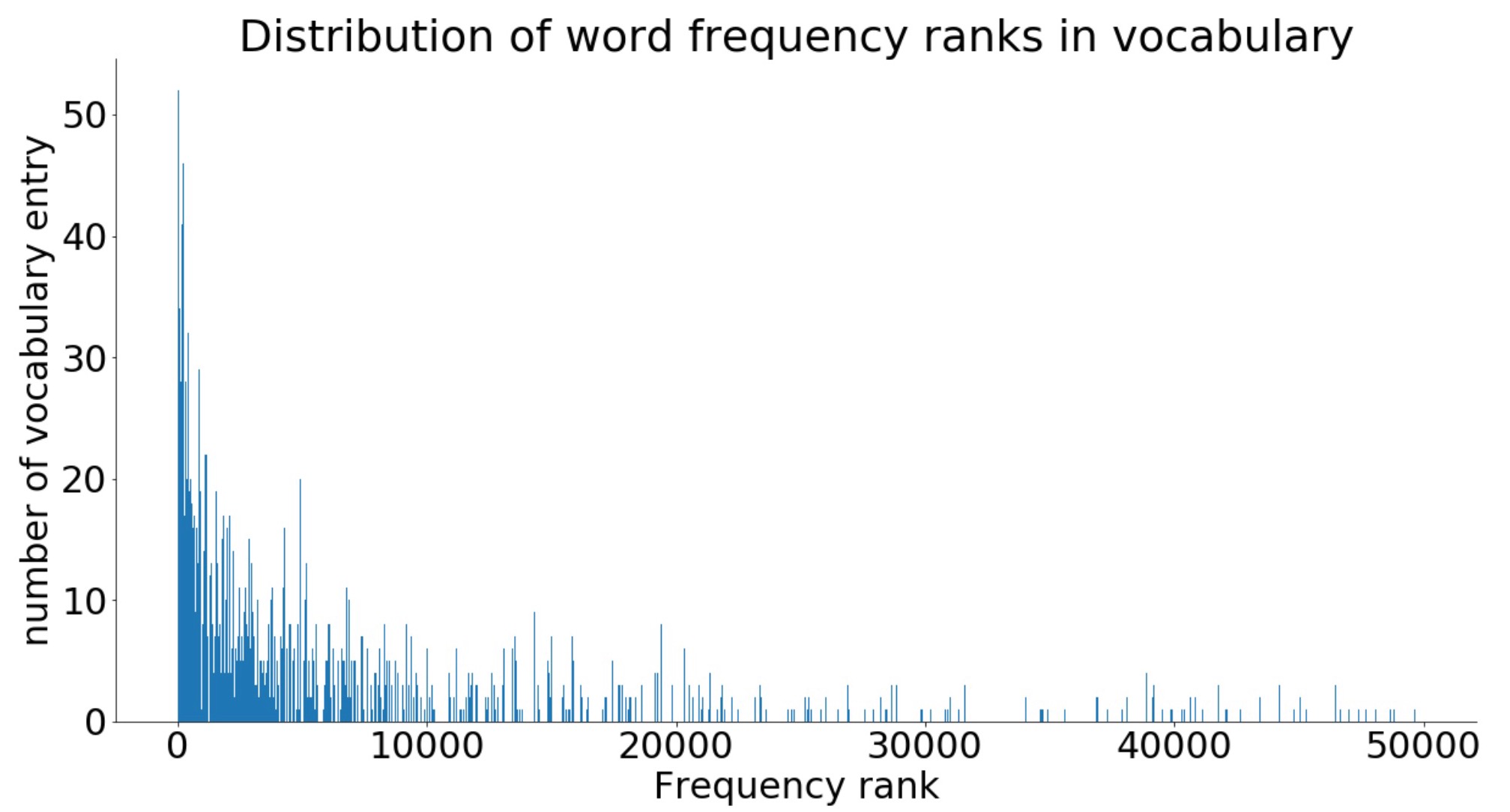}
     \caption{Comparison of word frequencies in CLiMP and the Chinese Internet Corpus.}
     \label{distribution}
 \end{figure}

We show the frequency of words in CLiMP's vocabulary in the Chinese Internet Corpus\footnote{http://corpus.leeds.ac.uk/frqc/internet-zh.num}  in Figure \ref{distribution}.
1,055 of the words in CLiMP are within the 5,000 most frequent words in the Chinese Internet Corpus.

\begin{table*}[t]
\centering
\newcommand{\tabincell}[2]{\begin{tabular}  
{@{}#1@{}}#2\end{tabular}} 
\small
\resizebox{\textwidth}{!}{%
\begin{tabular}{llll}
\toprule
\textbf{Phenomenon} & \textbf{N}& 
\textbf{Acceptable Example} & \textbf{Unacceptable Example}\\
\midrule

\tabincell{l}{Anaphor\\agreement} & 1 
& \tabincell{l}{\small{王玉珍 \ \ 震惊-了\quad \underline{她自己}。}\\
\small{Jane.F \quad shock-PST \ \underline{herself}.}\\
\emph{\small{'Jane shocked \underline{herself}.'}}} &\tabincell{l}{\small{王玉珍 \  \ 震惊-了\quad \underline{他自己}。}\\
\small{Jane.F \quad shock-PST \ \underline{himself}.}\\
\emph{\small{'Jane shocked \underline{himself}.'}}}\\

Binding & 1 
& \tabincell{l}{\small{杨颖 \ \ 治疗 吴宇涛 之后 \ 佩服-过 \quad \underline{她自己}。}\\
\small{Yang.F\ cure \ \  Wu.M\ \  after \ admire-PST \underline{herself}}\\
\emph{\small{'Yang admired \underline{herself} after she cured Wu.'}}} 
& \tabincell{l}{\small{杨颖 \quad 治疗 吴宇涛 之后 \  \ 佩服-过 \qquad \underline{他自己}。}\\
\small{Yang.F \ \ cure \ \  Wu.M\ \   after \ admire-PST \ \ \underline{himself}}\\
\emph{\small{'Yang admired \underline{himself} after she cured Wu.'}}}  \\

\tabincell{l}{\emph{b\v{a}}\\construction} & 1 & \tabincell{l}{\small{王鑫 \qquad \underline{把} \ 自行车 \ 扔  \quad 了。}\\
\small{Wong.M \ \underline{BA} \ \ bike \ throw \ PST}\\
\emph{\small{'Wong threw away the bike.'}}} &
\tabincell{l}{\small{王鑫 \qquad \underline{被} \ \ 自行车 \ 扔 \quad  了 。}\\
\small{Wong.M \ \underline{PASS}  \ bike \ throw \ PST }\\
\emph{\small{'Wong \underline{was thrown away} by the bike.'}}}\\

Coverb & 3 & \tabincell{l}{\small{李文清 \quad \underline{乘} \ 卡车 \quad到达-了\quad 咖啡店。}\\
\small{Lee.M \quad \ \underline{ride} truck \ arrive-PST coffee shop}\\
\emph{\small{'Lee went to the coffee shop \underline{by} truck.'}}} &
\tabincell{l}{\small{李文清 \quad \underline{于} 卡车 \quad到达-了\quad 咖啡店。}\\
\small{Lee.M \quad \underline{at}\quad truck \ \ arrive-PST coffee shop}\\
\emph{\small{'Lee went to the coffee shop \underline{at} truck.'}}}\\

NP head finality & 1 & \tabincell{l}{\small{王梦\qquad 正在 \ 卖\ \underline{张红梅 \ 清洗-过-的}\quad 推车。}\\
\small{Wong.F PROG sell \underline{May.F clean-PRF-ADJ} trolley}\\
\emph{\small{‘Wong is selling the trolley \underline{that Mel has cleaned}.’}}} &
\tabincell{l}{\small{王梦\qquad 正在 \ 卖\ \ 推车 \quad \underline{张红梅 \ 清洗-过-的}。}\\
\small{Wong.F PROG sell trolley \underline{May.F clean-PRF-ADJ}}\\
\emph{\small{‘Wong is selling the trolley \underline{that Mel has cleaned}.’}}} \\

Classifier & 2 & \tabincell{l}{\small{张杰\quad 正在\ \ 穿过\ \ 一\qquad \qquad \underline{家}\qquad \qquad 艺术画廊}。\\

\small{Jay.M PROG pass one \underline{CL:INSTITUTION} art gallery}\\
\emph{\small{'Jay is passing through \underline{an} art gallery.'}}} & 
\tabincell{l}{\small{张杰\ 正在\ \ 穿过\ \ 一\qquad  \quad \underline{段}\qquad \quad 艺术画廊}。\\
\small{Jay.M PROG pass one \underline{CL:LENGTH} art gallery}\\
\emph{\small{'Jay is passing through \underline{an} art gallery.'}}}  \\

Filler gap & 1 & \tabincell{l}{\small{图书馆，\quad 我 \ 开车 去-过 \ \underline{这个地方}。}\\
\small{The library, \ \  I \ drive \  to-PRF \underline{this place}}\\
\emph{\small{‘The library, I have driven to \underline{this place}.’}}} & 
\tabincell{l}{\small{图书馆，\quad \ \ 我 \ 开车 去-过 \quad  \underline{博物馆}。}\\
\small{The library, \quad I \ drive \ to-PRF \underline{the museum}}\\
\emph{\small{‘The library, I have driven to \underline{ the museum}.’}}}  \\

Passive & 1  &\tabincell{l}{\small{这些\quad 患者 \quad 被\qquad \underline{转移}-了。}\\
\small{These patient PASS \underline{transfer}-PST}\\
\emph{\small{'These patients were \underline{transferred}.'}}}&
\tabincell{l}{\small{这些\quad 患者 \quad 被\quad \underline{下降}-了。}\\
\small{These patient PASS \ \underline{fall}-PST}\\
\emph{\small{'These patients were \underline{fell}.'}}}\\

\tabincell{l}{Verb\\complement}& 5 & \tabincell{l}{
\small{王慧 \qquad 的\quad 文章 \quad 吓 \qquad \underline{坏} \quad 了 \ \  包曼玉 。}\\
\small{Wong.F POSS article frighten \underline{badly} PST \ \ Bao.F.}\\
\emph{\small{'Wong's article frightened Bao \underline{badly}.'}}} & 
\tabincell{l}{
\small{王慧 \qquad 的\quad 文章 \quad 吓 \qquad \underline{开} \quad 了 \ \  包曼玉 。}\\
\small{Wong.F POSS article frighten \underline{openly} PST \ \ Bao.F.}\\
\emph{\small{'Wong's article frightened Bao \underline{openly}.'}}} \\

\bottomrule
\end{tabular}
}
\caption{\label{MP-examples}Nine Chinese linguistic phenomena covered by CLiMP with acceptable and unacceptable sentence examples. Minimal differences are underlined. The second line of each example shows a gloss, the third line is an English translation. N represents how many paradigms (each with 1,000 examples) are within each phenomena. }
\end{table*}

\subsection{Linguistic Phenomena}
\label{ssec:title-authors}
CLiMP covers 9 major linguistic phenomena in Mandarin Chinese, cf. Table \ref{MP-examples}. They are picked from a comprehensive Chinese grammar book by  \citet{po2015chinese}. Following \citeauthor{po2015chinese}'s discussion, we now explain the phenomena not present in English. The \textbf{\emph{b\v{a}} construction} is an SOV construction involving the particle \emph{b\v{a}}, which precedes the object and moves the object to a position before the main verb. It is only grammatical with a subset of transitive verbs. \textbf{Coverbs} are verb-like items that precede the main verb in a serial verb construction. They almost invariably have to be used in conjunction with other verbs in a sentence. They share some properties with prepositions, but are not syntactically interchangeable with them. 
\textbf{Classifiers} obligatorily appear with nouns when those are modified by numerals or adjectives. Mandarin has dozens of classifiers, and nouns select the classifier they combine with. \textbf{Verb complements} follow a verb, often expressing a result or manner of an event. Not all verbs can be used with all complements, making certain combinations ungrammatical. \textbf{NP head finality} is present in Mandarin noun phrases. The relative clause precedes noun phrases.

\subsection{Data Validation}
To verify whether the MPs in our dataset show clear contrasts, we conduct two rounds of human validation with 22 annotators. They are all native speakers of Chinese, 14 females and 8 males, whose ages range from 20 to 48. All of them have at least a high school degree.

In our first human validation, each human annotator is assigned a subset (100 MPs) of a paradigm. We let them perform the same forced-choice task as our models: decide for each MP which sentence seems more acceptable. We discard one paradigm, the coverb-direction paradigm, after this validation, because its human validation accuracy is below $85\%$. The average human agreement for the remaining paradigms is $95.8\%$.

In the second human validation, we sample 15 MPs from each of the remaining paradigm, resulting in a dataset consisting of 240 MPs. 16 annotators complete the same forced-choice task on this dataset. We count a MP as valid if more than half of the annotators agree with its label. The human agreement on this dataset is 97.1\%, showing that our data creation results in valid examples.

\subsection{Comparison with BLiMP}
BLiMP consists of 67 datasets, each containing 1,000 MPs and organized by phenomenon into 12 categories. CLiMP only contains 16 datasets due to the less inflectional nature of Mandarin Chinese. 3 phenomena are covered by both corpora: anaphor agreement, binding, and filler-gap. The human agreement for these three phenomena in BLiMP is $97.5\%$, $87.3\%$, and $86.9\%$, respectively. The corresponding accuracies in CLiMP are $94.5\%$, $99\%$, and $100\%$, respectively. The overall human agreement for BLiMP is $88.6\%$, which is $7.2\%$ lower than for CLiMP. 

\section{Models and Methods}
We use accuracy for evaluation. A MP in CLiMP is classified correctly if a LM assigns a higher probability to the grammatical sentence than to the ungrammatical one. We evaluate statistical and neural LMs, including masked LMs. Corpora which contain 0.4M, 2M, and 21.5M sentences are used for further exploration. 
We also investigate the effect of  different tokenizations.\footnote{We use character tokenization and word tokenization (\href{https://github.com/fxsjy/jieba}{https://github.com/fxsjy/jieba}).}

\textbf{Chinese BERT }
BERT \citep{devlin-etal-2019-bert} is a transformer-based neural model \citep{Vaswani2017AttentionIA}. Here, we evaluate Chinese BERT.\footnote{
    \href{https://github.com/google-research/bert/blob/master/multilingual.md}{https://github.com/google-research/bert/blob/master/multilingual.md}
    }
This model has 12 layers, 768 hidden units, 12 attention heads, and 110M parameters. The training dataset contains 25M sentences. We assign probabilities to sentences with this model by masking the words in a sentence one by one, computing the probability of each masked word, and, finally, multiplying the probabilities of all words \citep{wang-cho-2019-bert,salazar-etal-2020-masked}.\footnote{\href{https://github.com/xu-song/bert-as-language-model}{https://github.com/xu-song/bert-as-language-model}}

\textbf{LSTM LMs }
We further evaluate 6 LSTM \citep{hochreiter1997long} LMs. These model have 2 layers, 200 hidden units, and 2 attention heads. We train them using Pytorch's word language model code\footnote{\href{https://github.com/pytorch/examples/tree/master/word\_language\_model}{https://github.com/pytorch/examples/tree/master/ word\_language\_model}} 
on 3 differently-sized Chinese Wikipedia corpora: 0.4M, 2M, and 21.5M sentences. We further compare word-level and character-level models (cf. Table \ref{human-model-results}). For evaluation, we employ code adapted by \citet{TACL2013} from \citet{gulordava-etal-2018-colorless}.\footnote{\href{https://github.com/sheng-fu/colorlessgreenRNNs}{https://github.com/sheng-fu/colorlessgreenRNNs}}

\textbf{\textit{n}-gram LMs }
Finally, we experiment with 4 different 5-gram LMs, which have been trained on 0.4M and 2M sentences from Chinese Wikipedia. For each corpus size, we train one word-based and one character-based LM. Those models are implemented using KenLM.\footnote{\href{https://kheafield.com/code/kenlm/}{https://kheafield.com/code/kenlm/}}

\begin{table*}
\centering
\footnotesize
\setlength{\tabcolsep}{5.pt}
\begin{tabular}{lllllllllll}
\toprule
 \textbf{Model} & \textbf{Overall} & \textbf{Clsfr.} & \textbf{V.Cp.} & \textbf{Hd.Fi.} & \textbf{The ba.} & \textbf{Coverb} & \textbf{Ana.Agr.} & \textbf{Pass.} & \textbf{Bind.}  & \textbf{Fi.Gap}\\
\midrule \it
Human &\it95.8 & \it99.7 & \it96.0 & \it100.0 & \it85.0 & \it92.5 & \it94.5 &\it 91.0 &\it99.0 &\it100.0 \\
Chinese BERT  & 81.8& 92.9 &\bf93.0 &53.1&69.0 &87.9&86.2&67.7 &50.8 &62.4 \\
LSTM-21.5M-word &62.8 &75.7 &74.0 &\bf81.4 &10.0 &47.0 &63.1 &68.4 &50.1 &41.5 \\
LSTM-21.5M-char &60.7 &56.1 &64.9&\bf89.1 &32.1 & 43.2& 57.0 &67.9 & 50.0 &68.8 \\
LSTM-2M-word & 66.0 &\bf77.8 & 73.8 &75.0 & 48.4& 43.4& 67.0& 68.0&50.0 &59.2\\
LSTM-2M-char & 60.4 &68.4 &68.1 &\bf86.3 &29.0 &28.5 &68.1 &68.4 &50.1&61.9\\
LSTM-2M-word &  60.6 &69.9 & 65.4&70.3&41.1 &38.8 &66.3 &\bf72.7 &50.0&55.2\\
LSTM-2M-char & 63.2  & 68.9& 69.7&\bf83.9 & 25.0& 45.6&67.7 & 74.3 &50.0&64.4\\


\emph{5}-gram-2M-word &59.0 &70.1 &71 &55.2&15.6&39.2&67.7&\bf72.0 &49.6 &40.0 \\
\emph{5}-gram-2M-char &65.7 & 70.6& \bf78.8&68.3 & 30.6& 53.9&65.8 & 64.8& 51.6&57.3 \\
\emph{5}-gram-0.4M-word &55.9 &66.4 &69.5 &46.3 &6.0 &37.0 &69.1&\bf77.8 &49.1 &25.2 \\
\emph{5}-gram--0.4M-char &60.0 &\bf71.5 &65.4 &70.5 &19.3 &46.5 &68.8 &68.7 &50.2 &48.4\\

\bottomrule
\end{tabular}%
\caption{\label{human-model-results}
Percentage accuracy of all humans and models on CLiMP. Random guessing yields an accuracy of 50\%. Bold numbers indicate the phenomenon each model is best at. Numbers in model names (21.5, 2, 0.4) refer to the number of sentences in the training corpus.
}
\end{table*}

\section{Results}
All results are shown in Table \ref{human-model-results}.

\textbf{Phenomenon-specific Results }
Our LMs perform best on classifier--noun agreement and verb complement selection: Chinese BERT's accuracy is only $6.8\%$ and, respectively, $3\%$ lower than that of humans on these two phenomena. LSTMs and \emph{5}-grams remain around $30\%$ behind humans, but still perform better on these phenomena than on others in CLiMP. This indicates that  Chinese LMs acquire local selection knowledge better than the linguistic knowledge needed to master other phenomena.

Our LMs stuggle most with the \emph{b\v{a}} construction, binding, and filler-gap dependencies.
All models perform close to chance level for binding, suggesting that they lack the hierarchical knowledge necessary to correctly resolve the structural relationship between a reflexive and its binder. Similarly, most models perform near chance on filler-gap dependencies. This suggests that they do not robustly represent long-distance dependencies.\footnote{A caveats applies: because Mandarin lacks \emph{wh}-movement, we test filler-gap dependencies using a topicalization construction more common in speech, and less likely to appear in the training corpora.} 

On the head-final construction, Chinese BERT performs surprisingly poorly as compared to the other models: only $53.1\%$ accuracy as compared to an average accuracy of $81\%$ by the LSTMs. The coverb construction, in contrast, is easy for Chinese BERT: it achieves $87.9\%$ accuracy, while the highest accuracy among all other models is $47\%$.

\textbf{Model-specific Results }
Comparing across models, Chinese BERT achieves by far the highest overall accuracy with 81.8\%.  
Our different LSTMs all perform worse, but obtain surprisingly similar scores: from $60.4\%$ to $66.0\%$. The performances of our \emph{5}-grams range from $55.9\%$ to $65.7\%$. Keeping tokenization and corpus size constant, three out of four \emph{5}-grams are outperformed by LSTMs. Thus, we overall find that neural models have advantages as compared to statistical models.

Comparing among the LSTMs, we find similarly to \citet{hu2020systematic} that the corpus size does not have much influence on the overall performance, with the caveat that these models perform close to chance. In contrast, a larger corpus size does result in a better performance in \emph{5}-grams. We also compare the effect of different tokenizations: Character-based \emph{5}-grams demonstrate better performance than word-based ones. For LSTMs, however, using characters only results in a better performance for our smallest corpus size (0.4M).

Compared to English LMs \cite{TACL2013}, the human–model gap is much bigger for Chinese models. While neither models nor datasets are directly comparable between our and previous work, this still suggests that more analyses and developments are needed for non-English models.



\section{Conclusion}
We introduced CLiMP, a suite of diagnostic test sets aimed at evaluating which syntactic phenomena Chinese LMs learn, and used it to evaluate 11 different models.
All LMs appeared to have learned local selectional restrictions, but struggled with argument structure alternations, hierarchical structure, and long-distance dependencies.
Chinese BERT performed best on CLiMP overall. However, it obtained a 14\% lower accuracy than humans, suggesting there is still much room for improvement. We hope that CLiMP will serve as a linguistically informed resource for benchmarking and analyzing future progress on Chinese LMs. CLiMP is available at \url{https://nala-cub.github.io/resources}.


\section{Acknowledgments}
We would like to thank the students from CU Boulder's CSCI/LING5832 in Spring 2020 for their feedback on this research. We are also grateful for the feedback of the anonymous reviewers.

\bibliography{anthology,eacl2021}
\bibliographystyle{acl_natbib}

\appendix

\end{CJK}
\end{document}